\documentclass[10pt]{article} % For LaTeX2e

\usepackage[preprint]{rlj} % Should be uncommented for preprint versions

\usepackage{amssymb}            % Defines common symbols like \mathbb R
\usepackage{mathtools}          % Extends amsmath, providing common math tools
\usepackage{mathrsfs}           % Enables \mathscr, which can work in cases that \mathcal does not
\usepackage{graphicx}           % For including images
\graphicspath{ {./figs/} }
\usepackage{subcaption}         % Allows for the use of subfigures and subcaptions
\usepackage[space]{grffile}     % For spaces in image names
\usepackage{url}                % For displaying URLs
\usepackage{lipsum}             % For placeholder text
\usepackage{xcolor}
\usepackage{physics}
\usepackage{algorithmic}
\usepackage[ruled]{algorithm2e}
\usepackage{mathtools}
\usepackage{ulem}
\newcommand{\compactpara}[1]{\noindent\textbf{#1}\quad}
\newcommand{\tsl}{\mathcal{T}^{\text{SL}}}
\newcommand{\tpe}{\mathcal{T}^{\text{PE}}}
\newcommand{\tuple}[1]{\langle #1 \rangle}
\newcommand{\tb}[1]{\textbf{#1}}

\newcommand{\fX}{\mathcal{X}}
\newcommand{\fY}{\mathcal{Y}}
\newcommand{\fS}{\mathcal{S}}

\newcommand{\fB}{\mathcal{B}}
\newcommand{\ns}{{|\fS|}}
\newcommand{\nb}{{|\fB|}}

\newcommand{\E}{\mathbb{E}}
\newcommand{\R}{\mathbb{R}}

\newcommand{\dfeat}{{d_\text{feat}}}

\newcommand{\dmodel}{{d_\text{model}}}
\newcommand{\attn}{\text{Attn}}

\newcommand{\mlp}{\text{MLP}}
\newcommand{\uniform}{\text{Uniform}}
\newcommand{\msve}{\text{MSVE}}
\title{Experience Replay Addresses Loss of Plasticity \\in Continual Learning}

\setrunningtitle{Experience Replay Addresses Loss of Plasticity in Continual Learning}

\author{Jiuqi Wang, \ Rohan Chandra, \ Shangtong Zhang}

\emails{jiuqi@email.virginia.edu, \ rohanchandra@virginia.edu, \ shangtong@virginia.edu}

\affiliations{
\textbf{Department of Computer Science, University of Virginia, USA}
}

\contribution{
    This paper proposes a new hypothesis that experience replay can address the loss of plasticity in continual learning.
    }
    {
    Experience replay has been adopted extensively by works in continual learning to counter catastrophic forgetting~\citep{lopezpas2017gem,chaudhry2021hindsight,riemer2018mer}. 
    We hypothesize that it can also address the loss of plasticity. 
    Notably, we do not claim any methodological contribution beyond our experiment design. 
    The readers should view this work as a hypothesis test.
    }

\contribution{
    We test our hypothesis on continual regression, classification and policy evaluation benchmarks and observe that Transformers using experience replay do not lose plasticity. 
    }
    {
    We adopt benchmarks from previous works.
    In particular, we use Slowly-Changing Regression~\citep{dohare2024lop} for regression, permuted MNIST for classification~\citep{kirkpatrick2017overcoming,dohare2024lop} and random Boyan's chain~\citep{boyan1999least,wang2025transformers} for policy evaluation.
    }

\contribution{
    We conjecture that the Transformers address the loss of plasticity with experience replay via in-context learning.
    }
    {
    We do not have sufficient evidence to conclude that in-context learning is the actual inner workings of our observation.
    It is merely one plausible explanation.
    }

 % Your keywords

\summary{We proposed a new hypothesis that experience replay can address the loss of plasticity in continual learning.
We tested our hypothesis on continual regression, classification and policy evaluation benchmarks and observed that by simply adding an experience replay and processing the data with Transformers, the loss of plasticity disappears.
We conjecture that Transformers were learning in context.}

\begin{document}

% \makeCover  % Create the cover page
\maketitle  % Make the title section

\begin{abstract}
    Loss of plasticity is one of the main challenges in continual learning with deep neural networks, where neural networks trained via backpropagation gradually lose their ability to adapt to new tasks and perform significantly worse than their freshly initialized counterparts. 
    The main contribution of this paper is to propose a new hypothesis that experience replay addresses the loss of plasticity in continual learning. 
    Here, experience replay is a form of memory. 
    We provide supporting evidence for this hypothesis. 
    In particular, we demonstrate in multiple different tasks, including regression, classification, and policy evaluation, that by simply adding an experience replay and processing the data in the experience replay with Transformers, the loss of plasticity disappears. 
    Notably, we do not alter any standard components of deep learning. 
    For example, we do not change backpropagation. We do not modify the activation functions. 
    And we do not use any regularization. 
    We conjecture that experience replay and Transformers can address the loss of plasticity because of the in-context learning phenomenon.
\end{abstract}

%%%%%%%%%%%%%%%%%%%%%%%%%%%%%%%%%%%%%%%%%%%%%%%%%%%%%%%%%%%%%%%%
%% Section: Introduction
%%%%%%%%%%%%%%%%%%%%%%%%%%%%%%%%%%%%%%%%%%%%%%%%%%%%%%%%%%%%%%%%
\section{Introduction}
\label{sec:intro}
Continual learning~\citep{thrun1998lifelong,parisi2019continual,wang2024surveycontinual}, or lifelong learning, describes a class of machine learning problems in which a learner learns from a long or endless stream of tasks.
Continual learning differs from learning on a single static task in that 
(i) the training data arrive individually or in small mini-batches; 
(ii) the learner can only observe each data point or mini-batch once in the course of learning;
(iii) there can be more than one task presented one after another
with the task boundary (i.e., the moment when the task changes) unknown to the learner.
Due to these characteristics, continual learning with neural networks often suffers from the loss of plasticity,
which refers to the phenomenon where the neural networks trained via backpropagation gradually lose their ability to learn new unseen tasks after training on previous tasks~\citep{lyle2023understanding, dohare2024lop}.
One well-accepted explanation behind this phenomenon is that more and more neurons become ``dead'', shrinking the model capacity as it fits on more tasks.   

The key contribution of this work is the proposal of a novel hypothesis that 
\emph{experience replay addresses the loss of plasticity in deep continual learning}, 
accompanied by supporting evidence. 
Experience replay is a form of memory and has been widely used in deep continual learning to tackle catastrophic forgetting --- another challenge present in this domain, 
where the learner forgets the knowledge gained from early tasks after training on new ones~\citep{lopezpas2017gem, rolnick2019experience,buzzega2020dark, chaudhry2021hindsight}. 
Continual learners with experience replay maintain a buffer named the replay buffer that stores experience in the past. The replay buffer usually has a fixed size and contains only a tiny portion of the history.
The learners utilize the data in the buffer to combat the loss of previously learned knowledge in deep neural networks and continuously update their buffer.

The most surprising result we observed in this paper is that the loss of plasticity vanishes by simply augmenting a continual learner with experience replay and using proper neural network architectures to process the data in the experience replay. 
This observation is consistent across a wide range of continual learning tasks, including continual regression tasks such as the  Slowly-Changing Regression benchmark~\citep{dohare2024lop}, continual classification tasks such as the permuted MNIST problem~\citep{kirkpatrick2017overcoming,dohare2024lop}, and the continual policy evaluation tasks such as Boyan's chain~\citep{boyan1999least}. 
Another surprising result we observed is that only the Transfomer~\citep{vaswani2017attention} neural network architecture is free from the loss of plasticity in our experiments. 
Other popular network architectures, such as multi-layer perceptrons (MLPs) and recurrent neural networks (RNNs), remain suffering from the loss of plasticity even with experience replay.

Notably, we do not alter any standard components of deep learning at all in our empirical study. 
By contrast, a large body of existing methods for mitigating the loss of plasticity need to periodically inject plasticity into the neural network by altering the backpropagation mechanism~\citep{nikishin2023deep,dohare2024lop, elsayed2024addressing}, use dedicated activation functions~\citep{abbas2023loss}, or add special regularizations~\citep{kumar2024maintaining}.
That being said, \emph{we do not claim that experience replay with Transformers can more effectively mitigate the loss of plasticity than previous methods}. 
Frankly, we do not experiment with any previous approaches at all. 
Instead, we argue that our hypothesis is interesting mainly because experience replay with Transformers is the least intrusive method to the standard deep learning practices and is thus potentially applicable to a wide range of scenarios.

Although we do not have decisive evidence, we conjecture that the observed success of experience replay with Transformers results from in-context learning \citep{brown2020incontext,laskin2022context}. 
In-context learning refers to the neural network's capability to learn from the context in the input during its forward propagation without parameter updates~\citep{dong2024survey,moeini2025survey}. 
Evidence testifies that expressive models, such as the Transformers, can implement some machine learning algorithms in their forward pass~\citep{vonoswald2023transformers,ahn2024transformers,wang2025transformers}. 
We speculate that the Transformers in our experiments are learning to implement some \emph{algorithm} in the forward pass, and being free from the loss of plasticity is a consequence of learning an algorithm, though it is not clear why learning an algorithm can resolve this issue. 
We may also potentially attribute the failures of the RNN and MLP in leveraging experience replay to their incompetence to learn in context, aligning with the current wisdom of the in-context learning community.

\section{Background}
We now elaborate on the task formulations and define the neural network architectures we use in our experiments.

\compactpara{Continual Supervised Learning} We use $\tsl$ to denote the set of supervised learning tasks the continual learner can encounter.
We employ $\tau = \qty(\fX, \fY, P_{XY})$ to characterize each $\tau \in \tsl$, where $\fX$ and $\fY$ denote the feature space and the target space, respectively, and $P_{XY}$ defines a joint distribution over $\fX \times \fY$. We assume that $\fX = \fX'$ and $\fY = \fY'$ for all pairs of tasks $\tau = (\fX, \fY, P_{XY}) \in \tsl$ and $\tau' = (\fX', \fY', P'_{XY})$.
This condition ensures that the shapes of the features and targets are consistent throughout the learning process.
Let $\Delta\qty(\tsl)$ be a distribution defined over $\tsl$.
In a continual supervised learning problem involving a sequence of $K$ tasks $\tau_1, \tau_2, \dots, \tau_K$, where each task is independently sampled from $\Delta\qty(\tsl)$ one after another, $\tau_{i+1}$ only begins upon the termination of $\tau_i$.
Let $P^{i}_{XY}$ denote the joint distribution $P_{XY}$ of the $i$-th task.
At task $\tau_i$, a total of $Nb$ pairs of $(x, y)$ will be independently sampled from $P^{i}_{XY}$, forming a sequence of training data $(x_1, y_1), (x_2, y_2), \dots (x_{Nb}, y_{Nb})$ grouped in $N$ mini-batches of size $b$.
The continual learner observes each mini-batch only once in the same order the data are generated.
In addition, the learner cannot access $K$ or $N$, implying that it does not know the number of tasks in the queue or the task boundaries.
The objective of the continual learner is to find a parametric function, $f_\theta: \fX \to \fY$ that minimizes $\E_{(x, y) \sim P_{XY}^i}\qty[\mathcal{L}(f_\theta(x), y)]$ for some loss metric $\mathcal{L}$ defined over $\fY \times \fY$ for each $\tau_i$.

\compactpara{Continual Policy Evaluation}Let $\tpe$ denote the set of policy evaluation tasks.
Each task $\tau \in \tpe$ consists of a Markov Reward Process (MRP,~\citet{sutton2018reinforcement}) $\tuple{\fS, p_0, P_S, r, \gamma}$ and a feature function $\phi: \fS \to \R^d$.
Here, $\fS$ denotes the state space, $p_0$ defines the initial state distribution, 
such that $S_0 \sim p_0$, $P_S$ is the state transition probability function, 
where $S_{t+1} \sim P_S(S_t)$, $r: \fS \to \R$ is a bounded reward function,
and $\gamma \in [0, 1)$ is a discount factor.
Similar to the supervised learning setting, we assume there exists a distribution $\Delta\qty(\tpe)$ over the policy evaluation tasks, and a sequence of $K$ independent tasks $\tau_1, \tau_2, \dots, \tau_K$ is drawn accordingly.
We employ $\tuple{\fS^i, p_0^i, P_S^i, r^i, \gamma^i}$ and $\phi^i$ to denote the MRP and the feature function of $\tau_i$, the $i$-th task in the sequence.
The MRP is unrolled to generate trajectory $S_0, R_1, S_1, R_2, S_2, \dots, S_{N-1}, R_N, S_N$, where $S_0 \sim p_0^i$, $S_{t+1} \sim P^i_S(S_t)$, $R_{t+1} \doteq r^i(S_t)$.
At each time instance $t$, the learner observes a transition $\qty(S_t, R_{t+1}, S_{t+1})$. 
Same as the supervised learning case, the learner is unaware of the task boundaries or the total number of tasks.
For policy evaluation tasks, one typically wishes to approximate the value function,
defined as
$
    v(s) \doteq \E\qty[\sum_{t=0}^\infty \gamma^t r(S_t) \Big| S_0 = s].
$
Suppose the Markov chain defined by $P_S^i$ is ergodic for all tasks $\tau_i, i = 1, \dots, K$.
Then, each Markov chain has a well-defined unique stationary distribution, denoted as $\mu_i$ for $\tau_i$.
Let $v^i$ be the true value function of the MRP of $\tau_i$. 
The learner needs to learn a parametric function $f_\theta: \R^d \to \R$, such that $f_{\theta}(\phi^i(s)) \approx v^i(s)$ for all $s \in \fS^i$.
We measure the approximation error by the mean square value error (MSVE), defined as
$
    \msve(\hat{v}^i, v^i) \doteq \E_{s \sim \mu^i} \qty[\qty(\hat{v}^i(s) - v^i(s))^2],
$
where $\hat{v}^i(s) \doteq f_\theta\qty(\phi^i(s))$.

\compactpara{Experience Replay}
A continual learner with experience replay keeps the last $n$ training examples in a replay buffer $\fB$ of size $n$ in a first-in, last-out (FILO) manner.
Each instance is a feature-label pair $(x, y)$ for continual supervised learning tasks and a transition tuple $\qty(\phi(s), r(s), \phi(s'))$ for continual policy evaluation tasks.

\compactpara{Transformer}
We employ a simplified version of Transformer made purely of attention layers~\citep{vaswani2017attention}.
Given a sequence of $n$ tokens in $\R^{\dmodel}$ as input,
a self-attention layer with weight matrices $W_k \in \R^{d_k \times \dmodel}, W_q \in \R^{d_k \times \dmodel}, W_v \in \R^{\dmodel\times\dmodel}$ processes the input $Z\in \R^{\dmodel \times n}$ as
$
    \attn_{W_k, W_q, W_v}(Z) \doteq W_v Z \sigma \qty(Z^\top W_k^\top W_q Z),
$
where $\sigma$ is the row-wise softmax activation. 
An $L$-layer Transformer is parameterized by
$
    \theta = \qty{\qty(W_k^{(i)}, W_q^{(i)}, W_v^{(i)})}_{i = 0, \dots, L-1}.
$
Denoting the input sequence as $Z^{(0)}$,
the matrix evolves layer by layer as 
$
    Z^{(l+1)} \doteq Z^{(l)} +  \attn_{W_k^{(l)}, W_q^{(l)}, W_v^{(l)}}\qty(Z^{(l)}).
$

Our Transformer lacks fully connected feed-forward layers, positional encoding~\citep{vaswani2017attention}, layer norms~\citep{ba2016layernorm} or even an output layer.
However, we observed that this stripped-down version of Transformer works well in practice and allows us to demonstrate the algorithmic power of attention and Transformer without the distraction from the commonly added components.

\compactpara{RNN}
We use the canonical RNN architecture with a hidden state initialized to zero and omit the parameterization details. 
Given a sequence of embeddings, the RNN evolves the hidden states embedding by embedding and layer by layer.

\compactpara{MLP}
We adopt a canonical MLP consisting of an input layer, a stack of hidden layers, and an output layer and apply an activation function after each layer except for the output.

\section{Addressing Loss of Plasticity in Regression}
We adopted a variant of the Slowly-Changing Regression problem~\citep{dohare2024lop} to study continual regression.
Slowly-Changing Regression is a simple benchmark that allows for fast implementation and experiments.
Furthermore, its simplicity does not compromise efficacy.
As we shall demonstrate later, loss of plasticity emerges even with a fairly shallow network, aligning with the observations by~\citet{dohare2024lop}.
These properties make Slowly-Changing Regression an ideal testbed for studying continual regression.
We leave the details of the task generation to supplementary material~\ref{supp: scr task generation}.

We first reproduced the loss of plasticity phenomenon by training an MLP.
We trained the model for 1000 tasks with $N=10,000$  and $b = 1$.
In other words, there were a total of 1000 flips where each flip happens after 10,000 pairs of data arrived one after another.
We performed one step of gradient descent on the squared loss $\qty(\mlp_\theta(\tilde x) - \tilde y)^2$ for each pair of data $(\tilde x, \tilde y)$ using the AdamW optimizer~\citep{loshchilov2018decoupled}.

The following experiments investigated the effectiveness of experience replay in addressing the loss of plasticity on Slowly-Changing Regression with a Tranformer, RNN and MLP learner.
We used the identical data presented in the same order as the previous experiment to make fair comparisons.
The learners equipped themselves with a replay buffer $\fB$ with a maximum size of 100.
Given the current training example $(\tilde x, \tilde y)$, we need to combine it with the memory stored in $\fB$ as input to our models.
Suppose $\fB$ is full and $\fB = \qty{\qty(x^{(i)}, y^{(i)})}_{i=1}^{100}$, we constructed embedding $Z$ as
\begin{align}
    Z = \mqty[x^{(1)} & x^{(2)} & \cdots & x^{(100)} & \tilde x\\
              y^{(1)} & y^{(2)} & \cdots & y^{(100)} & 0].
\end{align}
When $\nb < 100$ at the beginning of training, we set the first $100 - \nb$ columns of $Z$ to zero to keep the shape of the embeddings consistent.
The embedding $Z$ straightforwardly merges the memory and the input feature in one elongated matrix and is directly compatible with the Transformer and RNN.
The Transformer learner takes in this $Z$ and produces another embedding.
We read the last element of the last column of the output embedding as the approximated value of the learner.
The RNN learner processes $Z$ column-by-column and layer-by-layer, evolving its hidden states.
We take the last hidden state of the final layer of the RNN and extract its last element as the learner's output.
We are also curious if a simple neural network like an MLP can leverage experience replay to combat the loss of plasticity.
However, MLPs cannot process matrix inputs.
To this end, we flattened the embeddings column-wise into $\mqty[x^{(1)^\top} & y^{(1)^\top} & \cdots & x^{(100)^\top} & y^{(100)^\top} & \tilde x & 0 ]^\top$
before passing them to the MLP.
We name the MLP that consumes the memory as ERMLP (\tb{E}xperience \tb{R}eplay \tb{MLP})
to distinguish it from the one that only accepts $\tilde x$ as in the previous experiment.

At each time step, we constructed the embedding $Z$ as described above and minimized the squared loss $\qty(f_\theta(Z) - \tilde y)^2$ with an AdamW optimizer, where $f_\theta$ was implemented by a Transformer, an RNN, and an ERMLP, respectively.
We selected the depths and widths of the models such that their parameter counts were similar to the MLP\footnote {ERMLP was an exception due to the lengthy input vector.}.
After one gradient descent step, $(\tilde x, \tilde y)$ was pushed into $\fB$.
Thus, $\fB$ always kept the 100 most recent pairs.

\begin{figure}[ht]
    \includegraphics[width=\textwidth]{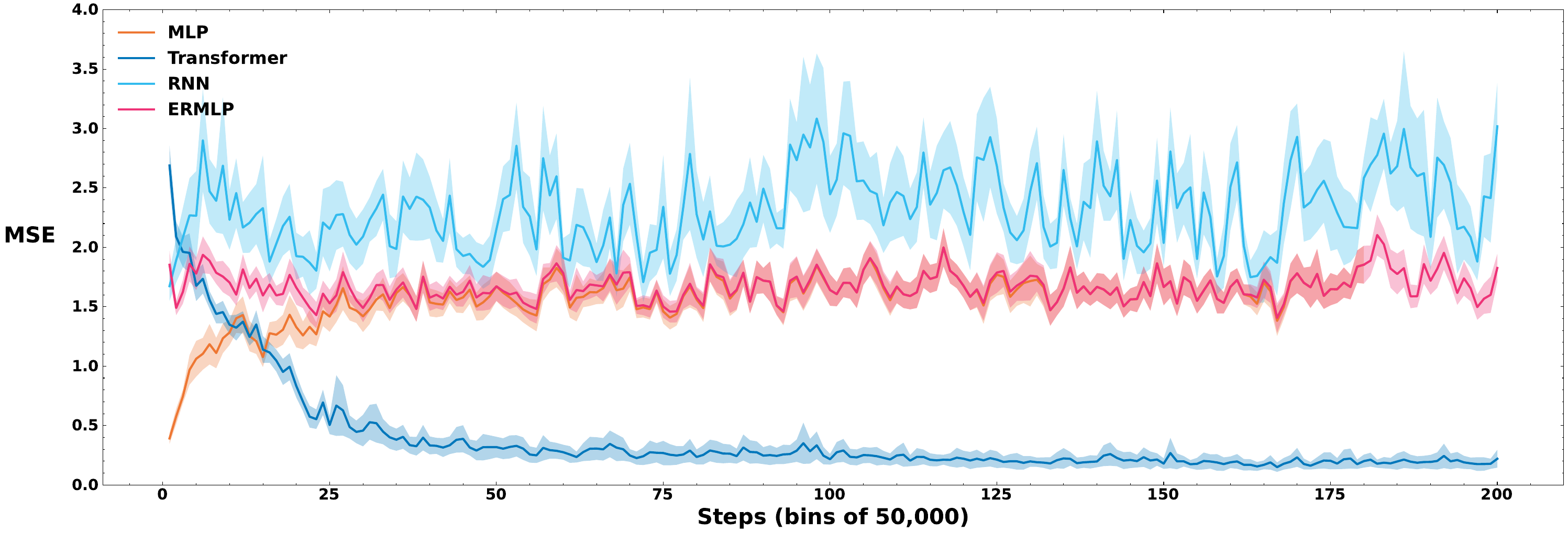}
    \caption{Mean square error for Slowly-Changing Regression. The losses are averaged in bins of 50,000. The runs are averaged over 20 seeds, and the shaded area displays the standard error.}
    \label{fig:reg loss}
\end{figure}

We plot the squared losses in Figure~\ref{fig:reg loss}, where every 50,000 losses are averaged to improve presentation.
The MLP lost plasticity rapidly and never restored, reflected by the rising then locally oscillating loss curve.
On the other hand, the Transformer did not suffer from the loss of plasticity, confirmed by the running loss.
What is more surprising is that the Transformer's performance increased as it trained on more tasks and did not degrade.
The RNN and ERMLP showed no clear signs of learning.
Therefore, the results suggest that the MLP does suffer from loss of plasticity as expected, whereas the Transformer is capable of leveraging memory to counter it.
The RNN and ERMLP fail to learn anything meaningful from the embedding in continual regression.
We include the details of this study in supplementary material~\ref{supp: scr config detail} for reproducibility.

\section{Addressing Loss of Plasticity in Classification}
To study continual classification, we employed permuted MNIST~\citep{kirkpatrick2017overcoming,dohare2024lop}, a widely adopted benchmark for continual classification.
MNIST~\citep{lecun1998mnist} is a non-trivial yet computationally cheaper dataset where a simple MLP can work reasonably well.
The permuted MNIST dataset builds on it by randomly shuffling the pixels of the images in the same way, allowing us to generate a virtually unlimited number of unique tasks for continual classification.
We put the details of the permuted MNIST task generation in supplementary material~\ref{supp: permuted MNIST generation}.

Similar to the regression experiment, we first demonstrate the loss of plasticity by training an MLP with three hidden layers of 2,000 units --- same configuration as~\citet{dohare2024lop}.
We trained the model for 7,000 tasks with $N = 150$ and a mini-batch $b=400$ due to computational limit.
Each $(\tilde x , \tilde y)$ pair in the mini-batch consists of an image flattened row-wise into a 49-dimensional vector and a label in one-hot representation in $\qty{0, 1}^{10}$.
The MLP took in $\tilde x$ and output a 10-dimensional vector $\mqty[z_1 & z_2 & \dots & z_{10}]^\top$ as the logits.
We applied softmax to normalize them into a probability distribution $\mqty[p_1 & p_2 & \dots & p_{10}]^\top$, where $p_k \doteq \exp(z_k)/\sum_{j = 1}^{10} \exp(z_j)$.
We performed one step of gradient descent to the cross-entropy loss
$
 -\frac{1}{b}\sum_{i=1}^b \sum_{k=1}^{10} y_k^{(i)}\log p_k^{(i)}
$
over the mini-batch using the AdamW optimizer.

We then repeated the permuted MNIST experiment with a Transformer, leveraging  experience replay.
We again presented the identical data as experienced by the MLP in the same order to the Transformer learner.
The learner kept a replay buffer $\fB = \qty{\qty(x^{(i)}, y^{(i)})}_{i=1}^{100}$ of size 100 used for constructing the embedding the same way as in the Slowly-Changing Regression experiment.
% We left-padded the embedding the same way as in the Slowly-Changing Regression experiment when $\fB$ was unfilled.
We read the last 10 elements of the last column of the embedding produced by the Transformer as its output logits.
We minimized the cross-entropy loss for the mini-batch using the same optimizer as the MLP.
Note that the memory used to construct $Z$ was constant within the same mini-batch, and we pushed the mini-batch into $\fB$ only after the gradient descent step.
Since the replay buffer was smaller than the mini-batch, we only retained the last 100 training pairs in the batch.

\begin{figure}[ht]
    \includegraphics[width=\textwidth]{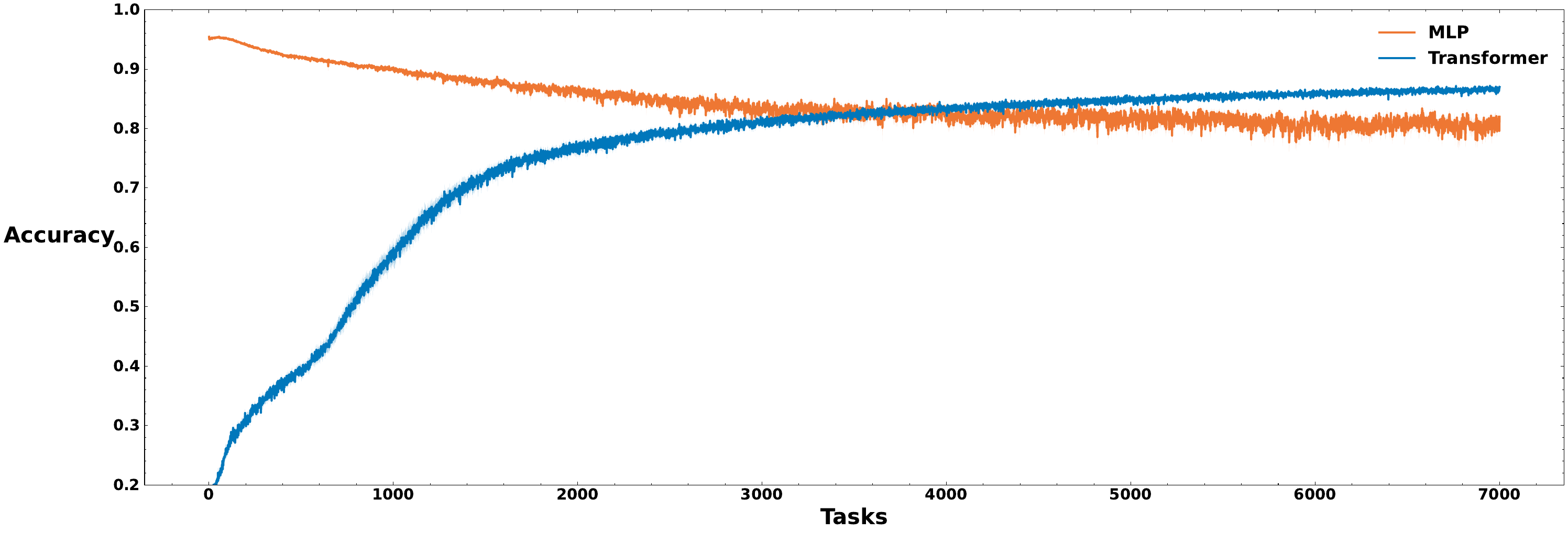}
    \caption{Test accuracy for permuted MNIST. The runs are averaged over 20 seeds, and the shaded area displays the standard error.}
    \label{fig:mnist acc}
\end{figure}

At the end of each task, we tested the models on the test set to obtain the prediction accuracies.
Note that when testing the Transformer, we used the latest replay buffer representing the most recent 100 pairs in the training set to form $Z$.
The buffer did not change throughout the testing phase for each task.
Figure~\ref{fig:mnist acc} shows the test accuracies.
The MLP again suffered from the loss of plasticity with a consistent decrease in test accuracies, aligning with the observations of~\citet{dohare2024lop}.
On the contrary, the Transformer improved its prediction accuracy by experiencing more tasks and manifested no hints of degradation.
Thus, these findings are consistent with our observations in the Slowly-Changing Regression experiment.
In addition, the parameter count of the Transformer ($\approx 70,000$) was less than 1\% of that of the MLP ($\approx 8,000,000$), yet the Transformer achieved better performance in the long run.
The RNN and ERMLP were not examined due to the limit of our computation resources (the training of RNN and ERMLP with long sequential input is extremely slow) and the knowledge that they do not tend to work well in continual supervised learning, as justified by the Slowly-Changing Regression problem.
We leave the detailed configurations of the permuted MNIST experiment to supplementary material~\ref{supp: permuted MNIST config detail}.

\section{Addressing Loss of Plasticity in Policy Evaluation}
So far, we have verified the viability of applying experience replay to address the loss of plasticity in continual supervised learning problems and obtained some positive results.
Here, we test our hypothesis on a continual policy evaluation problem.
We selected Boyan's chain~\citep{boyan1999least} as our MRP class because it is simple to construct, and we can fully control it.
A Boyan's chain has a chain of states, where each state except for the last two has $\epsilon$ probability of transitioning to the next state, $1 - \epsilon$ probability of transitioning to the second next state, for some $0 < \epsilon < 1$, and zero probability of going elsewhere.
The second last state transitions to the last state deterministically, while the last state transitions to all with nonzero probabilities.
Figure~\ref{fig:boyan chain} illustrates an example of the Boyan's chain with $m$ states adapted from~\citet{wang2025transformers}.
Every Boyan's chain is ergodic under this construction and thus has a unique and well-defined stationary distribution $\mu$.
This stationary distribution is also analytically solvable for finite-state ergodic Markov chains.
Since the state space of Boyan's chain is finite, one can represent the feature function $\phi$ as a matrix $\Phi \in \R^{\ns \times \dfeat}$, where $\phi(s)$ returns the $s$-th row of $\Phi$.
The details of the policy evaluation tasks generation procedure is in supplementary material~\ref{supp: bc generation}.

We chose an MLP with two hidden layers of 30 units that map a feature to a value estimation to show the loss of plasticity in continual policy evaluation tasks.
The learning algorithm is semi-gradient TD (or TD for short,~\citet{sutton1988learning, sutton2018reinforcement}), where, given observed transition $(\tilde\phi, \tilde r, \tilde \phi')$ and learning rate $\alpha$, the learner updates its parameters $\theta$ as
$
    \theta \gets \theta + \alpha\qty(\tilde r + \gamma \mlp_\theta\qty(\tilde \phi') - \mlp_\theta\qty(\tilde \phi))\nabla_\theta \mlp_\theta\qty(\tilde \phi).
$
Note that TD is not gradient descent because the gradient of the bootstrapped target $\gamma\mlp_\theta\qty(\tilde \phi')$ is not considered.
This nature of TD differentiates policy evaluation from the previous supervised learning tasks. 
We trained the model for 5,000 unique tasks with $N = 500$ transitions and a mini-batch size of $b=1$.
The optimizer was again AdamW.
With the same data and order, we trained models that employ experience replay.
The models equipped themselves with a replay buffer $\fB = \qty{\qty(\phi^{(i)}, r^{(i)}, \phi'^{(i)})}_{i=1}^{100}$ of size 100 for constructing embeddings
\begin{align}
    Z = \mqty[\phi^{(1)} & \phi^{(2)} & \cdots & \phi^{(100)} & \tilde \phi\\
               \gamma\phi'^{(1)} & \gamma\phi'^{(2)} & \cdots & \gamma\phi'^{(100)} & 0\\
               r^{(1)} & r^{(2)} & \cdots & r^{(100)} & 0],
    Z' = \mqty[\phi^{(1)} & \phi^{(2)} & \cdots & \phi^{(100)} & \tilde \phi'\\
                \gamma\phi'^{(1)} & \gamma\phi'^{(2)} & \cdots & \gamma\phi'^{(100)} & 0\\
                r^{(1)} & r^{(2)} & \cdots & r^{(100)} & 0].
\end{align}
Both $Z$ and $Z'$ were left-padded with zero for an unfilled replay buffer at the beginning of training.
The learners update their parameters $\theta$ as
$
 \theta \gets \theta + \alpha\qty(\tilde r + \gamma f_\theta(Z') - f_\theta(Z))
 \nabla_\theta f_\theta(Z),
$
where $f_\theta$ was implemented by a Transformer, an RNN, and an ERMLP, respectively.
We read the value approximated by the Transformer from the last element of the output embedding's last column.
Analogously, we extracted the value learned by the RNN from the last element of the final hidden state evolved by its final layer.
We flattened $Z$ and $Z'$ before passing them to the ERMLP.
The transition $(\tilde \phi, \tilde r, \tilde \phi')$ was pushed into $\fB$ after performing one step of TD.

\begin{figure}[ht]
    \includegraphics[width=\textwidth]{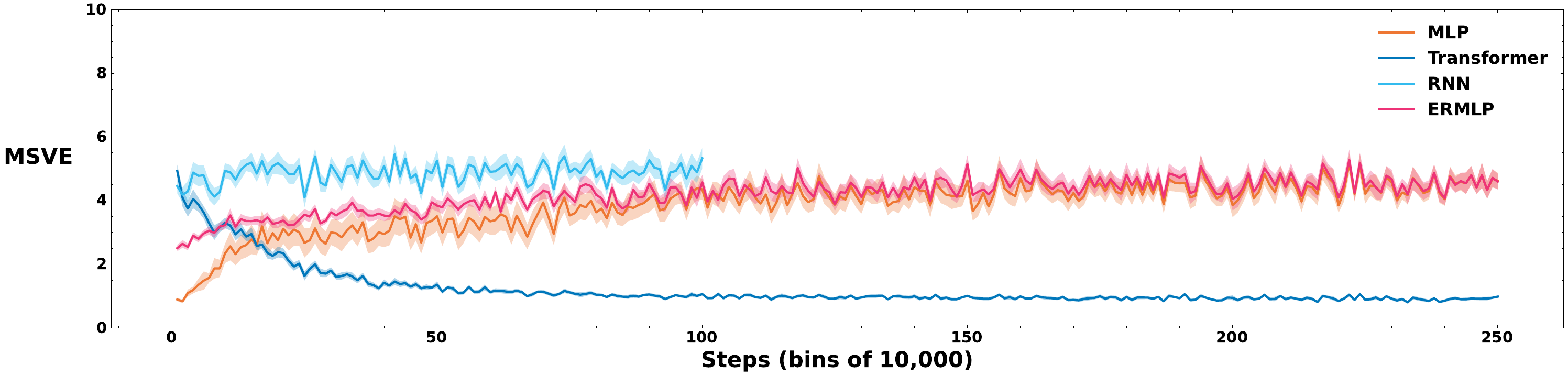}
    \caption{MSVE for policy evaluation with Boyan's chain. The MSVEs are averaged in bins of 10,000. The runs are averaged over 20 seeds, and the shaded area indicates the standard error.}
    \label{fig: pe msve}
\end{figure}

Since our Boyan's chain has a finite state space, and we have access to the stationary distribution and the true value function, the true MSVE is computable.
Hence, we recorded the MSVE of the models after every TD update.
The latest replay buffer content was used to construct the embeddings when testing models that leveraged experience replay.
Figure~\ref{fig: pe msve} plots the recorded MSVEs across tasks\footnote{Due to the sequential nature of the RNN, it trains significantly slower on our machine. We only have the results for the first 2,000 tasks, even though it has been running for over a week. The experiments are still running at the time of writing.}, averaged in bins of 10,000.
As expected, the prediction error of the MLP rose as it was trained on more tasks, signifying a loss of plasticity.
On the other hand, the Transformer gained performance with more tasks and did not lose plasticity at all.
These results again confirm that the loss of plasticity does emerge in continual policy evaluation problems as well, and using experience replay with a Transformer can address it.
Lastly, we did not observe any signs of learning in the RNN or the ERMLP for this task.
Readers can find the details of the continual policy evaluation experiment in supplementary material~\ref{supp: bc config}.

\section{Related Work}
\compactpara{Addressing Catastrophic Forgetting with Memory}
Experience replay has played a pivotal role in addressing catastrophic forgetting~\citep{mccloskey1989catastrophic} in continual learning.
Methods that employ a buffer to store a portion of the history to achieve balanced performance between new and old tasks are sometimes termed the replay-based~\citep{wang2024surveycontinual} or rehearsal-based~\citep{khetarpal2022towards} approaches.
GEM~\citep{lopezpas2017gem} represents an important work in this category that leverages experience replay to penalize the gradients that would increase the loss in the old tasks in continual classification tasks.
A-GEM~\citep{chaudhry2018efficient} is a well-accepted follow-up of GEM that ensures the mean episodic memory loss does not increase instead of considering all past tasks to save computation.
Another similar work is HAL~\citep{chaudhry2021hindsight} that learns ``anchor points'' using episodic memory to constrain the gradients.
Note that GEM, A-GEM and HAL assume the task boundaries are known because they require a task descriptor that distinguishes one task from another.
Rainbow Memory~\citep{bang2021rainbow} promotes diversity in episodic memory to improve performance under ``blurry'' task boundaries.
\citet{riemer2018mer} took a meta-learning approach to address catastrophic forgetting with experience replay in continual supervised and reinforcement learning tasks.
Another line of work directly lets the neural network rehearse on replay data.
\citet{hayes2019memory} maintained $K$ class-specific buffers of fixed size and trained the neural network on all buffers at each time step for continual image classification.
\citet{mai2020batch} mixed replay and online data at the batch level to rehearse.
CLEAR~\citep{rolnick2019experience} used experience replay to clone the old policies and penalize differences in value functions in continual reinforcement learning.

\compactpara{Addressing Loss of Plasticity}
\citet{lyle2023understanding} studied the mechanisms of the loss of plasticity and recommended network parameterizations and optimization choices less prone to this phenomenon.
The seminal work of~\citet{dohare2024lop} systematically investigated the loss of plasticity in continual supervised and reinforcement learning.
They proposed continual backpropagation that reinitializes inactive units periodically as a mitigation.
\citet{sokar2023dormant} studied the dormant neuron phenomenon in deep reinforcement learning and proposed an analogous method called ReDo that detects and reinitializes dormant units.
\citet{abbas2023loss} focused on the loss of plasticity in continual reinforcement learning exclusively and used Concatenated ReLU activations~\citep{shang2016understanding} as a remedy.
\citet{nikishin2023deep} proposed to address the loss of plasticity via plasticity injection.
UPGD~\citep{elsayed2024addressing} modified gradient descent to retain plasticity in the network.
\citet{kumar2024maintaining} alleviated the loss of plasticity via regularization.

\section{Discussion}
Based on our empirical studies, experience replay with Transformers can successfully address the loss of plasticity in continual supervised and reinforcement learning problems.
One may wonder what the underlying mechanism behind this remarkable phenomenon is. 
To this end, we conjecture that the improvements in the Transformers result from the emergence of in-context learning. 
In-context learning~\citep{brown2020incontext,laskin2022context,dong2024survey,moeini2025survey} refers to the neural networks' capability to learn new tasks from the data provided in the input in the forward pass without updating parameters. 
Existing literature shows that Transformers can solve regression~\citep{vonoswald2023transformers,ahn2024transformers}, classification~\citep{shen2025training} and policy evaluation~\citep{wang2025transformers} tasks in context. 
If the Transformers are indeed learning in context, they do not have to update their parameters for a new task in principle and thus remain plastic. 
In addition, compared with Transformers, works investigating in-context learning with MLP or RNN are much more scarce. 
\citet{wang2025transformers} demonstrate that an RNN likely cannot implement TD in its forward pass, partially explaining what we have seen in the experiments. 
Although our observations closely align with the in-context learning phenomenon, we do not draw a decisive conclusion that it is what is happening in the models.
At this stage, it is merely one possible explanation of the mechanism, and we leave the decoding to future work.

While this work is the first to verify the hypothesis that experience replay helps address the loss of plasticity to the best of our knowledge, there are several limitations.
First, our problems are simple in scale, leaving the scalability of this approach unexamined.
Second, processing long sequences with attention is expensive, as the complexity is $\mathcal{O}(n^2)$ for a sequence length of $n$ in attention layers.
The learners with experience replay run considerably slower than the MLP baselines, even with a memory of size 100.
One promising future extension is to use the structured state-space models~\citep{gu2021combining,gu2022efficiently,gu2024mamba} to speed up processing.
Thirdly, as previously discussed, we do not have an explanation or theoretical insights for this observation.
We believe addressing these limitations makes fruitful directions for future research.

\section{Conclusion}
This work proposes and tests the hypothesis that experience replay can address the loss of plasticity in neural networks.
We reproduce the loss of plasticity phenomenon with an MLP in continual regression, classification, and policy evaluation tasks. 
We also demonstrate that a Transformer does not suffer from this issue in all these problems with experience replay and gains performance as it trains on more tasks.
Furthermore, RNN and MLP fail to learn with experience replay.
We conjecture without proof that in-context learning emerges in the Transformers, allowing them to learn new tasks without losing plasticity.
Future research includes but is not limited to scaling the experiments to more complex problems, leveraging experience replay more efficiently, and finding explanations for this phenomenon.

\section*{Acknowledgments}
This work is supported in part by the US National Science Foundation (NSF) under grants III-2128019 and SLES-2331904. 

%%%%%%%%%%%%%%%%%%%%%%%%%%%%%%%%%%%%%%%%%%%%%%%%%%%%%%%%%%%%%%%%
%% Appendices
%%%%%%%%%%%%%%%%%%%%%%%%%%%%%%%%%%%%%%%%%%%%%%%%%%%%%%%%%%%%%%%%
\appendix

%%%%%%%%%%%%%%%%%%%%%%%%%%%%%%%%%%%%%%%%%%%%%%%%%%%%%%%%%%%%%%%%
%% NOTE: THIS MARKS THE END OF THE "MAIN TEXT"
%%%%%%%%%%%%%%%%%%%%%%%%%%%%%%%%%%%%%%%%%%%%%%%%%%%%%%%%%%%%%%%%

%%%%%%%%%%%%%%%%%%%%%%%%%%%%%%%%%%%%%%%%%%%%%%%%%%%%%%%%%%%%%%%%
%% Bibliography
%%%%%%%%%%%%%%%%%%%%%%%%%%%%%%%%%%%%%%%%%%%%%%%%%%%%%%%%%%%%%%%%
\bibliography{bibliography}
\bibliographystyle{rlj}

%%%%%%%%%%%%%%%%%%%%%%%%%%%%%%%%%%%%%%%%%%%%%%%%%%%%%%%%%%%%%%%%
% AUTHOR: If your paper has no supplementary materials, you may 
%         comment out the line below, which creates the title for
%         the supplementary materials.
%%%%%%%%%%%%%%%%%%%%%%%%%%%%%%%%%%%%%%%%%%%%%%%%%%%%%%%%%%%%%%%%
\beginSupplementaryMaterials

\section{Slowly-Changing Regression}
\subsection{Task Generation}
\label{supp: scr task generation}
Suppose the feature dimension is $m$, we initialize the first task $\tau_1$ by sampling a binary feature vector from $\qty{0,1}^m$ uniform randomly.
The first $f$ bits of the features are held constant throughout $\tau_1$, whereas the remaining $m-f$ bits are uniformly randomly flipped for each instance drawn from $\tau_1$.
Then, for each subsequent task $\tau_i$, the first $f$ bits of the features are obtained by randomly flipping one of the first $f$ bits of the features in $\tau_{i-1}$ and held constant during $\tau_i$.
The remaining $m-f$ bits are again randomly flipped at each step.
The targets are generated by passing the features through a fixed neural network having a hidden layer with the linear threshold unit (LTU) activation introduced in~\citet{dohare2024lop}.
The weights of the target-generating network are initialized randomly to be -1 or 1 at the beginning of training.
Under this protocol, the training data distribution shifts slightly after each task, requiring the learner to adapt continuously.
\subsection{Configuration Detail}
\label{supp: scr config detail}
\begin{table}[htbp]
    \begin{minipage}{0.45\textwidth}
    \centering
        \begin{tabular}{|c|c|}
           \hline
           \multicolumn{2}{|c|}{\tb{MLP}}\\
           \hline
           \# layers & 2\\
           \hline
           hidden dimension & 20\\
           \hline
           activation & ReLU\\
           \hline
           \hline
           \multicolumn{2}{|c|}{\tb{ERMLP}}\\
           \hline
           \# layers & 2\\
           \hline
           hidden dimension & 20\\
           \hline
           activation & ReLU\\
           \hline
           \hline
           \multicolumn{2}{|c|}{\tb{RNN}}\\
           \hline
           \# layers & 1\\
           \hline
           hidden dimension & 20\\
           \hline
           activation & tanh\\
           \hline
           \hline
           \multicolumn{2}{|c|}{\tb{Transformer}}\\
           \hline
           \# layers & 2\\
           \hline
           activation & softmax\\
           \hline
           \hline
           \multicolumn{2}{|c|}{\tb{LTU Net}}\\
           \hline
           \# layers & 1\\
           \hline
           hidden dimension & 100\\
           \hline
           activation & linear threshold\\
           \hline
           $\beta$ & 0.7\\
           \hline
        \end{tabular}
        \caption{Slowly-Changing Regression model configurations.}
    \end{minipage}
    \begin{minipage}{0.45\textwidth}
        \centering
        \begin{tabular}{|c|c|}
            \hline
            number of tasks & 1,000\\
            \hline
            instances per task & 10,000\\
            \hline
            feature dimension & 20\\
            \hline
            mini-batch size & 1\\
            \hline
            replay buffer capacity & 100\\
            \hline
            MLP learning rate & 0.01\\
            \hline
            ERMLP learning rate & 0.01\\
            \hline
            RNN learning rate & 0.001\\
            \hline
            Transformer learning rate & 0.0001\\
            \hline
            random seeds & 20\\
            \hline
        \end{tabular}
        \caption{Slowly-Changing Regression experiment parameters.}
    \end{minipage}
\end{table}
\section{Permuted MNIST}
\subsection{Task Generation}
\label{supp: permuted MNIST generation}
The MNIST dataset consists of 70,000 hand-written digits from 0 to 9 as $28 \times 28$ grayscale images and their labels, of which 60,000 are for training and 10,000 are in the test set.
Due to computational constraints, we downsampled the images to $7 \times 7$.
For each new task, we randomly generated a permutation of the 49 pixels and applied the same permutation to all 70,000 images.
The labels remained unchanged.
The continual learner must adjust to the new permutation once every 60,000 training pairs.
\subsection{Configuration Detail}
\label{supp: permuted MNIST config detail}
\begin{table}[htbp]
    \begin{minipage}{0.45\textwidth}
        \centering
        \begin{tabular}{|c|c|}
           \hline
           \multicolumn{2}{|c|}{\tb{MLP}}\\
           \hline
           \# layers & 3\\
           \hline
           hidden dimension & 2,000\\
           \hline
           activation & ReLU\\
           \hline
           \hline
           \multicolumn{2}{|c|}{\tb{Transformer}}\\
           \hline
           \# layers & 10\\
           \hline
           activation & softmax\\
           \hline
        \end{tabular}
        \caption{Permuted MNIST model configurations.}
    \end{minipage}
    \begin{minipage}{0.45\textwidth}
        \centering
        \begin{tabular}{|c|c|}
            \hline
            number of tasks & 7,000\\
            \hline
            training set size & 60,000\\
            \hline
            test set size & 10,000\\
            \hline
            image dimension & $7 \times 7$\\
            \hline
            mini-batch size & 400\\
            \hline
            replay buffer capacity & 100\\
            \hline
            MLP learning rate & 0.001\\
            \hline
            Transformer learning rate & 0.0005\\
            \hline
            random seeds & 20\\
            \hline
        \end{tabular}
        \caption{Permuted MNIST experiment parameters.}
    \end{minipage}
\end{table}

\section{Boyan's Chain}
\subsection{Task Generation}
\label{supp: bc generation}
For each task $\tau_i$, we sampled a 10-state Boyan's chain MRP and a feature function $\phi^i$ with $\dfeat=4$ following the procedure introduced in \citet{wang2025transformers} closely:
We randomly generated the transition probabilities $P_S^i$, retaining the chain's structure. 
Then, we sampled a 10-dimensional vector as the reward function $r^i$, where the $s$-th element was $r^i(s)$.
The feature function was generated by randomly sampling a feature matrix $\Phi^i \in \R^{10\times 4}$.
The only thing we did differently was setting the initial distribution $p_0^i$ to be the stationary distribution $\mu^i$.
Lastly, we set $\gamma = 0.9$.
We provide the details of this procedure in Algorithm~\ref{alg: pe generation} for completeness.
Suppose that, with a slight abuse of notation, $P_S^i$ is represented as a transition probability matrix in $[0,1]^{10 \times 10}$, 
where $P_S^i(s, s') \doteq \Pr\qty(S_{t+1} = s' \mid S_t = s)$, one can solve for the true value function $v^i$ using Bellman equation~\citep{sutton2018reinforcement} as $v^i = \qty(I - \gamma P_S^i)^{-1} r^i$.
Generally speaking, access to the true value function is impossible except for simple environments like Boyan's chain.
Knowledge of the true value functions allows us to evaluate our learners against the ground truths instead of estimations.

\begin{figure}[htbp]
    \includegraphics[width=\textwidth]{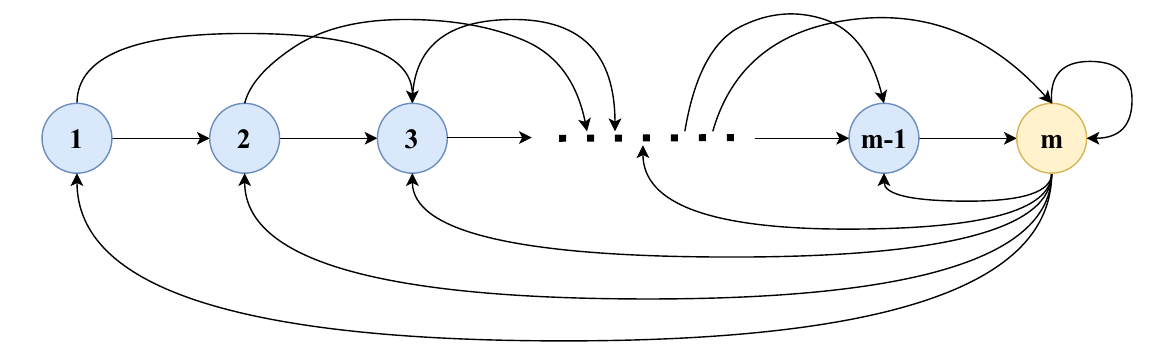}
    \caption{Boyan's chain example. Arrows indicate nonzero transition probabilities.}
    \label{fig:boyan chain}
\end{figure}

\begin{algorithm}
    \caption{Boyan's chain and feature generation 
    (adapted from~\citet{wang2025transformers})}
    \begin{algorithmic}[1]
        \label{alg: pe generation}
        \STATE \textbf{Input:} state space size $m = \abs{\fS}$, feature dimension $d$ 
        \FOR{$s \in \fS$}
        \STATE $\phi(s) \sim \uniform{(-1, 1)^d}$ \CommentSty{\, // feature}
        \ENDFOR
        \STATE $r \sim \uniform{(-1, 1)^m}$ \CommentSty{\, // reward function}
        \STATE $P \gets 0_{m \times m}$ \CommentSty{\, // transition matrix}
        \FOR{$i = 1, \dots, m - 2 $}
            \STATE $\epsilon \sim \uniform{(0, 1)}$
            \STATE $P(i, i+1) \gets \epsilon$
            \STATE $P(i, i+2) \gets 1 - \epsilon$
        \ENDFOR
        \STATE $P(m-1, m) \gets  1$
        \STATE $z \gets \uniform{(0, 1)^m}$
        \STATE $z \gets z/\sum_s z(s)$
        \STATE $P(m, 1:m) \gets z$ \CommentSty{\, // last state transition probabilities}
        \STATE $p_0 \gets \text{solve\_stationary\_distribution}(P)$
        \CommentSty{\, // initial distribution} 
        \STATE \textbf{Output:} $(p_0, P, r, \phi)$
    \end{algorithmic}
\end{algorithm}
\subsection{Configuration Detail}
\label{supp: bc config}
\begin{table}[htbp]
    \begin{minipage}{0.45\textwidth}
        \centering
        \begin{tabular}{|c|c|}
           \hline
           \multicolumn{2}{|c|}{\tb{MLP}}\\
           \hline
           \# layers & 2\\
           \hline
           hidden dimension & 30\\
           \hline
           activation & ReLU\\
           \hline
           \hline
           \multicolumn{2}{|c|}{\tb{ERMLP}}\\
           \hline
           \# layers & 2\\
           \hline
           hidden dimension & 30\\
           \hline
           activation & ReLU\\
           \hline
           \hline
           \multicolumn{2}{|c|}{\tb{RNN}}\\
           \hline
           \# layers & 6\\
           \hline
           hidden dimension & 9\\
           \hline
           activation & tanh\\
           \hline
           \hline
           \multicolumn{2}{|c|}{\tb{Transformer}}\\
           \hline
           \# layers & 6\\
           \hline
           activation & softmax\\
           \hline
        \end{tabular}
        \caption{Policy evaluation model configurations.}
    \end{minipage}
    \begin{minipage}{0.45\textwidth}
        \centering
        \begin{tabular}{|c|c|}
            \hline
            number of tasks & 5,000\\
            \hline
            instances per task & 500\\
            \hline
            number of states & 10\\
            \hline
            discount factor & 0.9\\
            \hline
            feature dimension & 4\\
            \hline
            mini-batch size & 1\\
            \hline
            replay buffer capacity & 100\\
            \hline
            MLP learning rate & 0.003\\
            \hline
            ERMLP learning rate & 0.003\\
            \hline
            RNN learning rate & 0.001\\
            \hline
            Transformer learning rate & 0.001\\
            \hline
            random seeds & 20\\
            \hline
        \end{tabular}
        \caption{Policy evaluation experiment parameters.}
    \end{minipage}
\end{table}
\end{document}